\documentclass{article}

\usepackage{microtype}
\usepackage{graphicx}
\usepackage{subfigure}
\usepackage{booktabs} 
\usepackage{colortbl} 
\usepackage{multirow}
\usepackage{hyperref}

\usepackage[accepted]{icml2025}

\usepackage{amsmath}
\usepackage{amssymb}
\usepackage{mathtools}
\usepackage{amsthm}
\usepackage{graphicx}
\usepackage{subcaption}
\usepackage[capitalize,noabbrev]{cleveref}

\definecolor{airforceblue}{rgb}{0.36, 0.54, 0.66}
\definecolor{lightblue}{rgb}{0.5, 0.8, 0.9} 
\usepackage{tcolorbox}
\newtcolorbox{mytextbox}[2][]{ 
    colback=lightblue!10,       
    colframe=lightblue!90!black,     
    fonttitle=\bfseries,        
    title=#2,                   
    coltitle=white,             
    sharp corners,              
    #1                          
}
\theoremstyle{plain}
\newtheorem{theorem}{Theorem}[section]

\newtheorem{lemma}[theorem]{Lemma}

\theoremstyle{definition}
\newtheorem{definition}[theorem]{Definition}

\theoremstyle{remark}

\DeclareMathOperator*{\argmin}{arg\,min}

\usepackage[textsize=tiny]{todonotes}

\icmltitlerunning{Submission and Formatting Instructions for ICML 2025}

\begin{document}

\twocolumn[
\icmltitle{ Data-Efficient Multi-Agent Spatial Planning with LLMs \\}












\icmlsetsymbol{equal}{*}

\begin{icmlauthorlist}
\icmlauthor{Huangyuan Su}{yyy,sch}
\icmlauthor{Aaron Walsman}{yyy,sch}
\icmlauthor{Daniel Garces }{sch}
\icmlauthor{Sham Kakade}{yyy,sch}
\icmlauthor{Stephanie Gil}{sch}
\end{icmlauthorlist}

\icmlaffiliation{yyy}{John A. Paulson School Of Engineering And Applied Sciences, Harvard University, 150 Western Ave, United States}
\icmlaffiliation{sch}{Kempner Institute, 150 Western Ave, United States}

\icmlcorrespondingauthor{Huangyuan Su}{csu@g.harvard.edu}
\icmlcorrespondingauthor{Stephanie Gil}{sgil@g.harvard.edu}

\icmlkeywords{Machine Learning, ICML}

\vskip 0.3in
]




\begin{abstract}


In this project, our goal is to determine how to leverage the world-knowledge of pretrained large language models for efficient and robust learning in multiagent decision making. We examine this in a taxi routing and assignment problem where agents must decide how to best pick up passengers in order to minimize overall waiting time. While this problem is situated on a graphical road network, we show that with the proper prompting zero-shot performance is quite strong on this task. Furthermore, with limited fine-tuning along with the one-at-a-time rollout algorithm for look ahead, LLMs can out-compete existing approaches with 50 times fewer environmental interactions. We also explore the benefits of various linguistic prompting approaches and show that including certain easy-to-compute information in the prompt significantly improves performance. Finally, we highlight the LLM’s built-in semantic understanding, showing its ability to adapt to environmental factors through simple prompts.
\end{abstract}

\section{Introduction}

Resource allocation in rapidly changing dynamic environments is an important logistics and controls problem with huge economic impact \cite{kondor2022cost, vazifeh2018addressing}.  However, many approaches to these problems train models for specific environments from scratch using model-based or model-free reinforcement learning (RL), which can be challenging in settings where data collection is expensive.
Recently, as large language models (LLMs) have increased in capability \cite{ahn2022can, o1}, they have been increasingly deployed on problems outside their traditional domain by developing prompting strategies that convert other data modalities into natural language.  In examples ranging from biology \cite{jumper2021highly} to chess \cite{zhang2024transcendence} to robotics \cite{ahn2022can, yang2023foundation}, researchers have shown that LLMs can accelerate learning and in some cases improve capabilities due to the inherent world knowledge that is built into them during pretraining.

In this work, we explore the application of LLMs to the challenging resource allocation problem of multi-agent taxi routing in a graph-based road network. Specifically, for each agent, we provide textual descriptions of the underlying environment, the current state of all agents, and all outstanding requests as input to the LLM. The LLM outputs its reasoning and final decision for where to travel next or pickup a passenger for that agent. We show that Llama3-8B-Instruct \cite{dubey2024llama} not only offers strong zero-shot performance in this setting, but also that when fine-tuned and combined with the rollout algorithm \cite{bertsekas2021multiagent, bertsekas1999rollout}, it can outperform previous state-of-the-art approaches with fewer total environment interactions during training.  This efficiency is important in real-world problems that are difficult to simulate and where data collection is expensive.


We also offer guidance on prompt engineering for graph-based data and demonstrate that incorporating easily computable information from the underlying physical environment—such as shortest paths to passengers—can significantly enhance performance. 

In a preliminary case study, we demonstrate the ability of LLMs to enhance planning performance by incorporating additional contextual information: they can react to semantic descriptions of the environment to anticipate relevant changes in dynamics. For example, an LLM prompted with the additional information that it is currently raining can predict increased cost of traveling and react accordingly. We include an example of this in the Appendix \cref{sec: semantic-info}.

Finally, we show that the use of LLMs in this setting is not without drawbacks. In some scenarios the LLM will hallucinate and provide controls that are impossible given the current graph structure and outstanding requests. We show that these can be largely mitigated by finetuning or incorporating feasibility checking-then-resampling.

Our primary contributions are:
\begin{enumerate}
\item Through experiments, we demonstrate that even in zero-shot settings, Llama 3-8B-Instruct performs remarkably well. Moreover, our finetuned rollout policy surpasses previous state-of-the-art approaches while requiring significantly less training data. Finetuning also effectively eliminates spatial hallucinations observed in the prompting-based methods.
\item We conduct a comprehensive ablation study on prompting techniques, offering valuable insights into how LLMs can serve as planners for multi-agent tasks.
\item We demonstrate that our prompting-based methods generalize effectively to a significantly larger map with more agents, featuring a quadratic increase in the number of nodes and roads.
\end{enumerate}




\section{Problem Formulation}

\begin{figure}[ht]  
    \centering  
    \includegraphics[width=0.6\columnwidth]{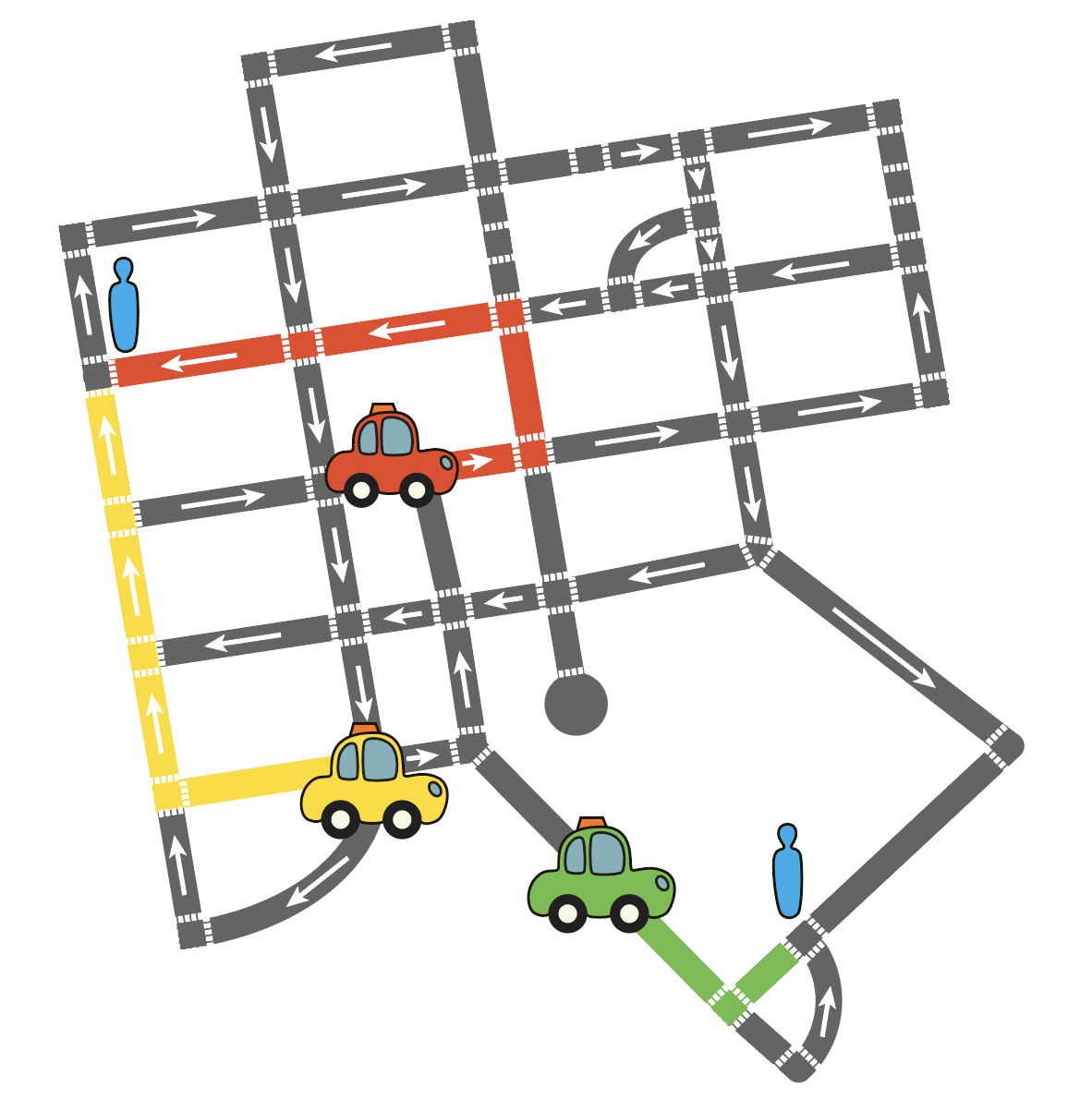}  
    \caption{In this multi-agent setting, taxis must decide where to move and pickup passengers in order to minimize overall passenger wait time. At each time step, taxis can move to any neighboring intersection, where arrows indicate one-way road segments.  In this example there are two outstanding requests, one in the top left and another in the bottom right.  While the green taxi is closest to the request in the bottom right, the task of allocating the red and yellow taxi is more challenging.  Both are four stops away from the request in the top left, but which one is sent, and where the other moves in the meantime will determine how prepared the taxis are for picking up future requests.}  
    \label{fig:fig_1}  
    \vskip -0.1in
\end{figure}

\subsection{Environment} We study the problem through a multi-agent taxicab routing problem introduced in \cite{garces2023multiagent}. Given an urban environment with a fixed street topology, represented as a directed graph, $\mathbb{G} = (\mathbb{V}, \mathbb{E})$, where $\mathbb{V} = [n]$ is the set of indices of the street intersections, and $\mathbb{E} \subseteq \{(i, j)|i, j \in \mathbb{V} \}$ is the set of directed streets connecting adjacent intersections. We define $\mathcal{N}(i) = \{j|j \in \mathbb{V},(i,j) \in \mathbb{E}\}$ as the neighboring intersections for intersection $i$.   

Requests stochastically enter the system according to unknown distributions. A \textbf{request} $r$ is defined by the closest intersections to its desired pickup and dropoff locations, the time at which it enters the system; and if it has been assigned to any agent.  We define a \textbf{scenario} as a fixed map topology and a sequence of requests.

\subsection{State and Action} We assume there are a total of $m$ agents and each agent can perfectly observe all requests, and agents’ locations and occupancy status. We use letters $s, u$ to denote numerical states and controls, and letters $x, y$ to denote language input and output sequences. The state $s_k$ at time $k$ contains locations for all $m$ agents in terms of node indices, the time remaining in the currently assigned trip for all agents, and the set of outstanding requests at time $k$. If the remaining time for an agent $l$ is zero, the agent is considered available, and new requests can be assigned to it. 

We define the action space for an agent $l$ at state $s_k$ as $U^l_k(s_k)$. At each time step, this action space contains the following actions. If an agent is not currently assigned, the agent may move to one of the neighboring intersections, remain in its current position, or pick up a passenger at the current location. On the other hand, if it is assigned, it must move along the shortest path to the passenger's current destination. 

Our formulation of the action space allows for separable control constraints for each agent. For this reason, we express the joint control space at time $k$, $U_k(s_k)$, as the Cartesian product of the local control sets $U^1_k(s_k) \times \cdots \times U^m_k (s_k)$. As the control space grows exponentially with the number of agents, finding an optimal policy becomes intractable, necessitating the search for suboptimal solutions. 

\subsection{Stochastic Dynamic Programming} This routing problem can be formulated as a finite horizon, stochastic Markov Decision Process (MDP). 
The global \textbf{objective} of all agents is to find a pickup strategy that minimizes the total wait time of passengers
starting from the initial state $s_1$. A policy \(\pi = \{\mu_1, \dots, \mu_N \}\) is a list of functions, where $\mu_k$ maps state $s_k$ into control \(u_k = \mu_k(x_k) \in U_k(x_k)\). We define the cost of a policy $\pi$ starting at state $s_1$ as:
\[J_{\pi}(s_1) = \mathbb{E} \left[g_N(s_N) + \sum_{k=1}^{N - 1} g_k(s_k, \mu_k(s_k))\right].\]

In the equation above, $s_{k+1}$ is the state at time $k+1$ after application of control $u_k$ at time $k$ from the current state $s_k$, and $g_k(s_k, u_k)$ is the per-step cost, which is the number of current outstanding requests.
Note that the agents' actions do not affect the distribution of incoming requests.

\section{Background and Preliminaries}

\subsection{(One-at-a-time) Rollout and Offline Approximation} 
\label{sec: rollout-offline}
\begin{figure}[ht]
    \centering
    \centerline{\includegraphics[width=\columnwidth]{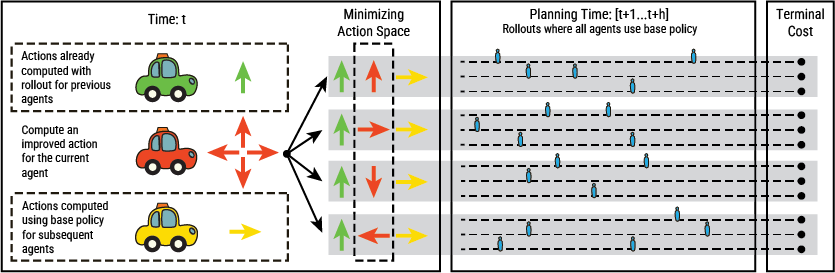}}
    \caption{We use multi-agent rollout \cite{bertsekas2021multiagent} for policy improvement. In this setting, we compute updated actions for each agent by computing many rollouts for each possible action it could take, and we choose the action resulting in the lowest average cost. In order to avoid computing rollouts for all combinations in the exponential joint multi-agent action space, agent actions are computed one at a time. In the example shown here, we are estimating an improved action for the red taxi. Each of its actions are combined with the best action found for the green taxi which has already been computed, and an action chosen by the base policy of the yellow taxi, which has not yet been updated by rollout.}
    \label{fig: rollout}
    \vskip -0.1in
\end{figure}

We adopt the \textbf{one-at-a-time rollout} framework from \cite{bertsekas2021multiagent}, which solves multiple smaller lookahead optimizations to derive a policy that improves upon an easy to compute heuristic known as the base policy. Agent $l$’s one-at-a-time rollout control is conditioned on the actions of other agents, where the controls for agents $1, \dots, l-1$  are already determined, and the actions for agents $l+1, \dots, m$ are computed using the base policy,  $i.e.$, .
\begin{align}
\label{eqn}
    \Tilde{u}_k \in \argmin_{u_k^l \in U_k^l(s_k)} \mathbb{E} \left[g_k(s_k, u_k) + \Tilde{J}_k(s_{k+1}) \right]
\end{align}

where \(u_k = (\Tilde{u}_k^1, \dots, \Tilde{u}_k^{l-1}, u_k^l, \mu_k(s_k), \dots, \mu_k(s_k) )\). We use \(\in\) instead of \(=\) as the optimal control is generally not unique. This process is illustrated in greater detail in Fig. \ref{fig: rollout}. 

Notably, this approach ensures that the control space scales linearly with the number of taxis, rather than exponentially. This is crucial given the high computational cost of LLM inference, as it avoids an exponential increase in inference calls. This rollout also allows for faster generation of high-quality training samples for policy evaluation. Furthermore, we use certainty equivalence \cite{bertsekas1999rollout} approximation to reduce the number of samples required to approximate the expectation, without changing the rollout algorithm. \noindent In our implementation of the CE, we fix the disturbances in distribution parameters across all the rollout steps, only preserving the stochasticity of the order in which requests arrive and the pairing between pickup and dropoff locations for each request.  

\subsection{Online Play and Approximate Policy Iteration}
Assuming our finetuned offline model successfully approximates the rollout policy, we then apply the offline trained models as base policy in online play, which according to recent theory \cite{bertsekas2022lessons} serves as an approximate Newton step to further improve the performance of the learned policy. Online play’s control $u_k$ at state $s_k$ is given by \cref{eqn}, with $\bar{u} = (u_k^1, \dots, \bar{u}^{l - 1}, u^l_k, \hat{\mu}(F(s_k, l+1)), \dots, \hat{\mu}(F(s_k, m)))$.

If the policy approximation, $\hat{\pi}$, correctly approximates the rollout policy $\tilde{\pi}$ on the current demand model, we expect online play policy $\bar{\pi}$ to outperform $\tilde{\pi}$ following the cost improvement property of the approximate policy iteration~\cite{bertsekas2019reinforcement}. If $\hat{\pi}$ fails to approximate the rollout policy $\tilde{\pi}$, the cost improvement property will not hold and the online play will not provide a significant improvement \cite{garces2023multiagent}.

Then, an \textbf{offline approximation} $\hat{\pi}$ of this one-at-a-time rollout-based RL policy is achieved by supervised finetuning of Llama 3-8B model. To train the policy approximation, we generate a dataset of random initial taxi and request locations. The training feature at state $s_k$ for an agent contain the state and other agent’s actions, \textit{i.e.}, $l \in \{1,...,m\}$ is $F(x_k, l) = (x_k, u_k^1, \dots, u_k^{l-1}, \mu^{l+1}_k(x_k), \dots, \mu^m_k(x_k))$ and the training label is $\mu_k^l$, which are the corresponding one-at-a-time rollout controls. 


\section{Our Method}

\subsection{Language Models as Base Policies}
\label{sec: prompt-methods} 
We study what physical information is pertinent to LLM decision making, how to encode that information in language for better LLM decision making, and how to use physical information to alleviate hallucinations~\cite{gil2023physicality}. 

We use \(p\) to denote a pre-trained LLM. For language model policies, we use chat template \cite{chat} to format the inputs and encapsulate the environmental information in the system prompt. An example of the system prompt is given in \cref{sec: sys_prompt}. For all policies that use LLMs, we use prompts to convert state and action information (of other agents) as inputs to the LLMs: \(x_k = \text{prompt}(s_k)\) and parse the outputs of the LLMs to get the actions: \(u_k = \text{parse}(y_k)\). The prompting function wraps input $x$ with task instructions. Furthermore, we physically ground the LLMs by also providing the shortest paths to outstanding requests, which is fast to compute from the environment. A concrete example is given in \cref{sec: cot_prompt}. We denote logically-coherent intermediate reasoning steps by \(z_1, \dots, z_n\). Furthermore, we define \textbf{hallucinations} as when the LLM outputs a next position that is neither a neighbor of the current position nor on the paths to one of the requests. In addition to zero-shot prompting, we also experiment with the following common techniques for improving performance.


\textbf{Few-shot.} 
We provide five examples consisting of input-output sequence pairs that illustrate the desired behavior across various states.

\textbf{Chain-of-Thoughts} \citep{wei2022chain} (CoT) enhances language models' ability to solve complex, multi-step problems by prompting the model to generate the reasoning steps explicitly. \(z_i \sim p^{CoT}(z_i| x, z_{1\dots i-1})\). \(y \sim p^{CoT}(y|x, z_{1\dots n})\) for  reasoning by encouraging step-by-step problem-solving and making the reasoning process more interpretable. An example of our CoT prompt is provided in \cref{sec: cot_prompt}. 

\textbf{Self-Consistency} \citep{wang2022self} \textbf{w/ CoT} (CoT-SC) samples $c$ independent chains of thought: \(\forall i\in [c]\), \([z^{(i)}_{1\dots n}, y^{(i)}] \sim p^{CoT}(y|x)\), and returns the most frequent output. It enhances CoT by exploring diverse thought processes for more reliable decisions. However, it lacks local exploration within chains, and its “most frequent” heuristic works best in limited output spaces.

\textbf{Tree-of-Thoughts} \cite{yao2024tree}(ToT). Instead of generating thoughts in a linear sequence, ToT explores a tree-like structure of possible thoughts or reasoning steps. Each node \(s^{\text{ToT}} =[x, z_{1\dots i}]\) represents a partial solution, allowing the model to identify the most promising global outcomes by breadth-first or depth-first search with lookahead and backtracking. $S^{\text{ToT}}$ is the state space of all such nodes. We design a value prompt \(V(p, S^{\text{ToT}}) \sim p^{value}(v|s^{\text{ToT}})\), \(s^{\text{ToT}} \in S^{\text{ToT}}\) to reason about the cost of taking an action. We restrict the tree depth to 1, making the search algorithm simply selects the branch with the highest value (or lowest cost). An example of our ToT prompt is provided in \cref{sec: tot_prompt}. 

\textbf{Zero-shot w/ Hallucination Checking }(ZS-HC). We design a checker that detects hallucinations based on the underlying environment and incorporates this information into the prompt for reprompting.




\section{Experiments}

\subsection{Implementation Details}

Our main models are zero-shot \textbf{Llama 3-8B} with/without rollout, finetuned \textbf{Llama 3-8B} with/without rollout and zero-shot \textbf{Llama 3.2-3B} \cite{llama32} with/without rollout. The Llama 3 models \cite{llama3} are state-of-the-art open-source large language models that improve upon the Llama 3 herd of models \cite{dubey2024llama} in terms of reasoning capabilities, context length, and tool use. The 8B and the 70B models can satisfy a large range of use cases. For our experiments, we use the 8B model because it has reasonable zero-shot performance on our problem and can be finetuned within reasonable time and infrastucture resources.

For fine-tuning, we adapt Llama Cookbook~\cite{llamarecipes}, the official codebase released for fine-tuning and building applications. Finetuning data is collected using rollout with an instantaneous reassignment policy (IA-RA, see Section \ref{subsec:baselines} for details) as the base policy and 2000 Monte Carlo (MC) simulations.  The details for finetuning using one-at-a-time rollout, are detailed in \cref{sec: rollout-offline}. We set the learning rate to \(5 \times 10^{-5}\), the batch size to \(8\), and adhered to the default settings in Llama Recipes for all other configurations. 

The Llama 3.2-3B models are expected to perform worse on this planning problem due to their optimization for multilingual dialogue tasks and their smaller sizes. Nonetheless, we use this smaller model as a baseline to study the trade-off between speed and performance.


\subsection{Baselines Used for Comparison}
\label{subsec:baselines}

We compare our approach with the following baselines:

\textbf{Greedy policy}: Each taxi moves towards its closest request without coordinating with other taxis. This method does not consider future demand.

\textbf{Instantaneous assignment (IA-RA)}: It matches available taxis and outstanding requests at every time step using an auction algorithm \cite{bertsekas1979distributed}. This method also does not consider future demand.

\textbf{Graph Neural Networks (GNNs) }\cite{garces2023multiagent}: Graph neural networks are an intuitive choice for this problem since the underlying environment is a road graph. Therefore, it serves as a good baseline that is of a different model architecture from transformers. For each load level, we train a separate pair of GNNs. The first GNN determines if an available agent should pick up a request in its current location, and the second GNN determines the next intersection towards which the agent should move. The move model is a multi-classifier while the pick up model is a binary classifier. 

\textbf{Rollout with Different Base Policies.} We also compare against rollout with greedy policy, IA-RA, and GNNs as the base policies. Rollout with GNNs as base policy is the previous state-of-the-art on this multi-agent routing problem.

\begin{figure*}[ht]
    \centering
    \centerline{\includegraphics[width=2\columnwidth]{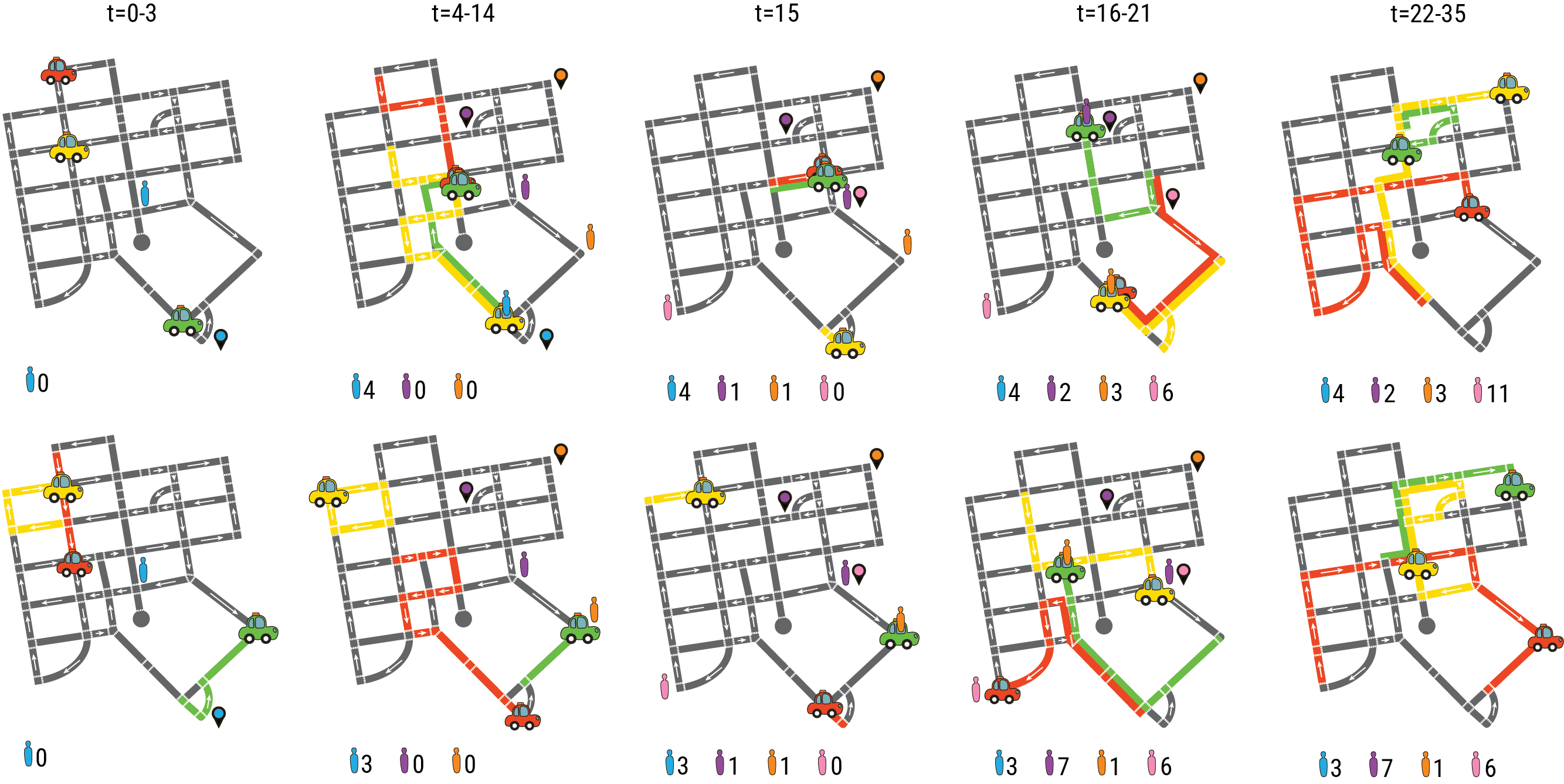}}
    \caption{Here we show two policies in the same scenario, greedy on the top row and Rollout using LLM as the base policy on the bottom.  Each column shows a time range, and the location of each taxi at the end of this range.  Colored lines show the route taken by the taxi during this interval.  In the first three frames on the far left, the LLM already starts to move the taxis into a more central location.  The first request (blue) arrives in the center of the map on frame 3, and the LLM picks it up one step earlier.  Eleven frames later on frame 14, two more requests (orange and purple) appear, then a third (pink) one frame later.  While the LLM is slower to pick up the purple request, it is able to pick up the orange and pink requests much faster and wins by three points over all.}
    \label{fig: }
\end{figure*}



\subsection{Hardware} We conduct experiments on Nvidia H100 or A100 GPUs. 
We use two techniques to enhance inference speed: (1) running all LLMs in FP8 precision, and (2) employing tensor parallelism across four GPUs within the same node, as these approaches provide results with negligible performance loss.

\subsection{Performance Benchmark}
We mainly consider a fixed horizon of \(60\) steps, which is equivalent to \(60\) minutes. We consider a \(400\times 400 m^2\) section of San Francisco with \(42\) nodes and \(125\) edges, and \(m = 3\) taxicabs. We consider 1-minute edge travel time. For rollout, we set the planning horizon to $t_h = 10$. 

We create a test set comprised of 20 scenarios. We create one set for each load level: low, medium, and high. The primary focus of our experiments is to evaluate the performance of the methods in terms of \textbf{cost}, defined as the total waiting time (in minutes) for all riders. Additionally, we report the occurrence of hallucinations alongside \textbf{cost}. Final performance is calculated as an average across all test scenarios.

\begin{table*}[t]
    \centering
  \caption{Performance comparison of methods in terms of total waiting time in minutes. We use {\color{lightblue!40} blue} to highlight the minimum cost and \textbf{bold} the minimum cost across all base policies. For \textbf{Llama 3-8B} zero-shot, we use the best number among all zero-shot prompting methods. $\dagger$: running with number of MC simulations set to 2000; $\ddagger$: running with number of MC simulations set to 200.}
    \label{tab:comparison}
    \resizebox{0.9\textwidth}{!}{
    \begin{tabular}{llcccccc}
        \toprule
        \multirow{2}{*}{\textbf{Method}} & \multirow{2}{*}{\textbf{Version}} & \multicolumn{2}{c}{\textbf{Low}} & \multicolumn{2}{c}{\textbf{Medium}} 
        &\multicolumn{2}{c}{\textbf{High}} \\
        \cmidrule(lr){3-4} \cmidrule(lr){5-6} \cmidrule(lr){7-8}
       &  & \textbf{Cost $\downarrow$} & \textbf{Hallucination  $\downarrow$} & \textbf{Cost $\downarrow$} & \textbf{Hallucination  $\downarrow$} & \textbf{Cost $\downarrow$} & \textbf{Hallucination  $\downarrow$} \\
             \midrule
       \rowcolor{gray!20}  \multicolumn{8}{c}{\textbf{Heuristic}}  \\
\multirow{2}{*}{\textbf{Greedy}} & Base & 10.16 & N.A. & 34.70 & N.A. & 61.28 & N.A. \\
  & Rollout$^\dagger$ & 8.95 & N.A. & 30.74 & N.A. & 52.39 & N.A. \\
\midrule
\multirow{2}{*}{\textbf{IA-RA}} & Base & 10.26 & N.A.\textbf{} & 31.78 & N.A. & 58.21 & N.A. \\
  & Rollout$^\dagger$ & 8.11 & N.A. & 27.22 & N.A. & 50.05 & N.A. \\
\midrule
\multirow{2}{*}{\textbf{Llama 3-8B }Zero-shot} & Base & $8.84$ & $3.95$  & 30.94 & 3.28 & 56.53 & 3.58  \\
  & Rollout$^\ddagger$ & 8.89 & 2.71 & 29.94 & ${2.91}$ & $56.63$ & $3.55$ \\
\midrule
\multirow{2}{*}{\textbf{Llama 3.2-3B }Zero-shot} & Base & $12.88$ & 3.47  & $37.88$ & $4.72$ & $65.89$ & $4.26$\\
  & Rollout$^\ddagger$ & $10.68$ & $3.44$ & $33.78$ & $4.93$ & $67.11$ & $5.06$ \\
               \midrule
       \rowcolor{gray!20}  \multicolumn{8}{c}{\textbf{Learning-based}}  \\
       \multirow{2}{*}{\textbf{GNN}} & Base & \textbf{8.37} & N.A. &  $30.04$ &  N.A. & $58.71$ & N.A. \\
  & Rollout$^\dagger$ & $8.23$ & N.A. & $27.38$ & N.A. & $47.32$ & N.A. \\
\midrule
\multirow{2}{*}{\textbf{Llama 3-8B} Finetuned} & Base & $8.68$ & 0  & \textbf{29.22} & 0  & $\mathbf{51.00}$ & 0 \\
  & Rollout$^\ddagger$ & \cellcolor{lightblue!40 }$7.05$ & 0 & 27.12\cellcolor{lightblue!40 } & 0 & \cellcolor{lightblue!40 }$47.11$ & 0\textbf{} \\
  
        \bottomrule
    \end{tabular}
    }
    \vskip -0.1in
\end{table*}




\subsection{Main Results}

We present our main results in \cref{tab:comparison} and discuss our observations and analysis below.

\textbf{Best Performing Model} Finetuning Llama, along with our offline and online play scheme outperforms all other models, including the previous state of the art GNN+rollout.  Importantly, unlike the GNN-based models, which trained a separate policy for each load level, we were able to finetune a single model for all load levels, suggesting better generalization.  Finally, the performance benefits of finetuning and rollout suggest that it is necessary to perform both the offline approximation and online play steps. While each pair of GNN models were trained with 6400 trajectories, we finetune a single Llama 3-8B model for all load levels with only 128 trajectories collected using Rollout with IA-RA as  base policy. With its larger capacity compared to GNN models, it demonstrates superior generalization capabilities. As a result, our approach achieves significantly greater \textbf{data efficiency} than the previous GNN-based method. 


We also found that even the zero-shot approach performs well, slightly trailing rollout with greedy or IA-RA base policies but outperforming greedy/IA-RA without rollout across all load levels. This may be due to its inherent behaviors: prioritizing high-index nodes and navigating to nearby nodes to maximize encounters with new requests.

Zero-shot methods can exhibit spatial and reachability hallucinations, leading to suboptimal behaviors (see \cref{sec: hallu}).  While cost grows as the level of load grows, this is not the case for the number of hallucinations. Furthermore, when comparing in a large map, we see that the number of hallucinations grows linearly with the number of agents. This suggests that the hallucination rate is an inherent property of the LLMs, which is not obviously affected by load levels in the environment.

\textbf{Rollout vs. Fine-tuning. } We also observed that, through only one round of finetuning, we are able to remove spatial hallucinations.  With the improvement of the cost and alleviating hallucinations, we conclude that finetuning is better than rollout with LLMs as base policies at improving the behavior of the LLMs.
When zero-shot prompted LLMs are used as the base-policy for rollout, we see that costs for rollout increase.  We hypothesize that this is caused by excess hallucinations during MC sampling, which biased the computed expected future cost. In other words, the LLM is not able to correctly approximate future costs and hence the rollout is not able to act as a Newton Step.


\subsection{Prompting Methods}
In this section, we present the results of applying various prompting techniques \cref{sec: prompt-methods} to Llama 3-8B, aiming to identify optimal zero-shot performance (without any training or fine-tuning). For zero-shot, CoT, and ToT, we set the temperature of the LLM to 0. While for self-consistency, we set the temperature to 0.7 as that technique requires diverse samples. For ZS-HC, we set the maximum number of reprompt to 5. 

\begin{table*}[t]
    \centering
  \caption{Performance comparison of prompting methods in terms of total waiting time in minutes. We \textbf{bold} the best number across all base policies and use {\color{lightblue!40} blue} to denote best number across all methods.}
    \label{tab:prompt}
    \resizebox{0.8\textwidth}{!}{
    \begin{tabular}{llcccccc}
        \toprule
        \multirow{2}{*}{\textbf{Method}} & \multirow{2}{*}{\textbf{Version}} & \multicolumn{2}{c}{\textbf{Low}} & \multicolumn{2}{c}{\textbf{Medium}} 
        &\multicolumn{2}{c}{\textbf{High}} \\
        \cmidrule(lr){3-4} \cmidrule(lr){5-6} \cmidrule(lr){7-8}
       &  & \textbf{Cost $\downarrow$} & \textbf{Hallucination  $\downarrow$} & \textbf{Cost $\downarrow$} & \textbf{Hallucination  $\downarrow$} & \textbf{Cost $\downarrow$} & \textbf{Hallucination  $\downarrow$} \\
             \midrule
       \rowcolor{gray!20}  \multicolumn{8}{c}{\textbf{}}  \\
{\textbf{Zero shot}} & Base & 8.84 & 4.37 & 32.75 & 3.28 &  56.53 & 3.58 \\
\midrule
{\textbf{Few shot}} & Base & 9.26 & 2.58 & 32.06 & 3.00 & 57.68 & 1.79 \\
\midrule
{\textbf{CoT-SC}} & Base & 9.37 & 2.21 & 32.83 & 2.50 &  55.84 & 2.72 \\
\midrule
{\textbf{CoT}} & Base & 8.68  & 2.32 & 30.06 \textbf{} & 3.94 & 56.95 & 3.24 \\
\midrule
{\textbf{ToT}} & Base & 8.63 & 1.79  & \cellcolor{lightblue!40}29.50 & 2.33 & 55.63 & \textbf{}2.11 \\
\midrule
{\textbf{ZS-HC}} & Base & \cellcolor{lightblue!40}8.42 & \textbf{1.67} & 30.89  & \textbf{1.35} & \cellcolor{lightblue!40 }55.05 & \textbf{1.32} \\
        \bottomrule
    \end{tabular}
    }
\end{table*}

We present the complete numerical results of all prompting methods in \cref{tab:prompt}. We observe that CoT, ToT have the effect of alleviating hallucinations and decreasing cost. The worse performance of ToT than CoT may be because that we ask the Llama 3-8B model to be the evaluator, which is too challenging for this model. 

However, from the results, we see that CoT and ToT offer only limited improvement of the total cost under heavy load levels. Therefore, we need better methods to improve LLMs' spatial reasoning abilities. We collect data using the simple zero-shot prompt, to ensure a fair comparison between prompting methods and finetuning/rollout methods.

Surprisingly, the results also indicate that few-shot and CoT-SC do not perform better than zero-shot prompting. 
This may be caused by the model making improper generalization (extrapolation) from the given examples.

\subsection{Ablations on Test-time Efficiency}
Compute consumption increases linearly with the number of MC simulations, which can be viewed as a hyperparameter of the rollout algorithm. This raises a critical question: how many MC simulations are necessary? To address this, we analyze the trade-off between computational cost and the number of MC simulations, using our fine-tuned 8B model. \cref{sample-complexity} shows that decreasing the number of MC simulations only leads to slightly worse performance. 

\begin{figure}[ht]
\begin{center}
\centerline{\includegraphics[width=\columnwidth]{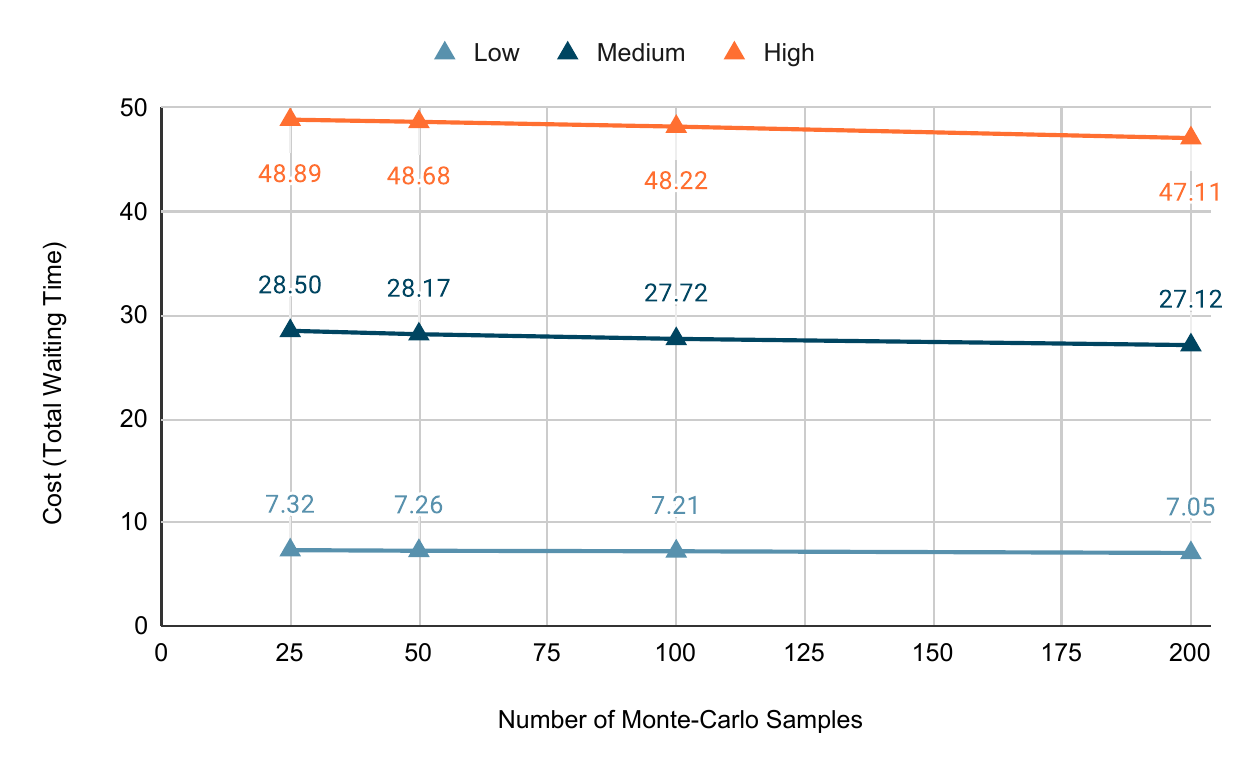}}
\caption{Cost (total waiting time) versus the number of MC futures sampled under low, medium, and high load levels.}
\label{sample-complexity}
\end{center}
\vskip -0.1in
\end{figure}
 \vspace{-10pt}
 
\subsection{Ablations on Scalability}
We study the scalability through two sets of experiments. First, we present performance on the most difficult scenario in our test in \cref{tab:hard}. We observe that the finetuned LLM and rollout with finetune LLM as the base policy are better in these hardest scenarios, suggesting that they are better at generalization. Second, we scale our environment to a $1000\times1000 m^2$ map and 10 agents in \cref{tab:large-map}. The numbers show that LLMs can scale to this setting.

\begin{table}[t]
    \centering
  \caption{Performance comparison of methods in the two hardest scenarios. Color coding is the same as in \cref{tab:comparison}.}
    \label{tab:hard}
    \resizebox{0.5\textwidth}{!}{
    \begin{tabular}{llccc}
        \toprule
        \multirow{2}{*}{\textbf{Method}} & \multirow{2}{*}{\textbf{Version}} & \multicolumn{2}{c}{\textbf{Cost}$\downarrow$}  \\
        \cmidrule(lr){3-4} 
       &  & Scenario 12 & Scenario 14    \\
             \midrule
             \multirow{2}{*}{\textbf{Greedy}} & Base &  172  & 121 \\
  & Rollout & 161 & 109  \\
             \midrule
             \multirow{2}{*}{\textbf{IA-RA}} & Base & 183 & 117 \\
  & Rollout & 160 & 115  \\
             \midrule
       \rowcolor{gray!20}  \multicolumn{5}{c}{\textbf{Prompting} Llama 3-8B}  \\
\multirow{2}{*}{\textbf{Zero-shot}} & Base & 187 & \textbf{105}  \\
  & Rollout & \textit{167} & \textit{108} &  \\
\midrule
\textbf{Few-shot} & Base & 188 & 113  \\
\midrule
\textbf{CoT}  & Base & 188 & {108} & \\
               \midrule
\textbf{CoT-SC} & Base & 163 & 108\textbf{} & \\
               \midrule
\textbf{ToT} & Base & 184 & 116 & \\
               \midrule
               \rowcolor{gray!20}  \multicolumn{5}{c}{\textbf{Finetuned/Trained}}  \\
       \multirow{2}{*}{\textbf{GNN}} & Base & 161 &  108\textbf{}  \\
  & Rollout & 159\textbf{} & \textbf{} 105  \\
  \midrule
    \multirow{2}{*}{\textbf{Llama 3-8B} Finetuned} & Base & \textbf{160} &  114 \\
  & Rollout &  151\cellcolor{lightblue!40 } & \textbf{}  \cellcolor{lightblue!40 }104  \\
        \bottomrule
    \end{tabular}
    }
    \vskip -0.1in
\end{table}


\begin{table}[t]
    \centering
  \caption{Performance comparison of methods in terms of total waiting time in the large map with more agents. Color coding is the same as in \cref{tab:comparison}. }
    \label{tab:large-map}
    \resizebox{0.5\textwidth}{!}{
    \begin{tabular}{llcccc}
        \toprule
        \multirow{2}{*}{\textbf{Method}} & \multirow{2}{*}{\textbf{Version}} & \multicolumn{2}{c}{\textbf{Low}} & \multicolumn{2}{c}{\textbf{Medium}} 
        \\
        \cmidrule(lr){3-4} \cmidrule(lr){5-6} 
       &  & \textbf{Cost $\downarrow$} & \textbf{Hallucination  $\downarrow$} & \textbf{Cost $\downarrow$} & \textbf{Hallucination  $\downarrow$}   \\
             \midrule
       \rowcolor{gray!20}  \multicolumn{6}{c}{\textbf{}}  \\
       \multirow{2}{*}{\textbf{Greedy}} & Base & 13.05 & N.A. & 37.40 & N.A.   \\
        & Rollout & 8.30 & N.A. & 25.05 & N.A.  \\
\midrule
\multirow{2}{*}{\textbf{IA-RA}} & Base & 12.45 & N.A. & 31.20 & N.A.   \\
 & Rollout & \cellcolor{lightblue!40 } 8.20 & N.A. & 24.70 & N.A. \\
\midrule
{\textbf{Zero-shot}} & Base & 10.30 & 12.95 & 27.15 & 20.80 \\
\midrule
{\textbf{CoT-SC}} & Base & \textbf{8.58} & 9.94 & \cellcolor{lightblue!40 } 
 \textbf{24.20} & 20.05  \\
        \bottomrule
    \end{tabular}
    }
    \vskip -0.1in
\end{table}




\section{Related Works}
\subsection{Multi-agent Decision-making}
Multi-agent decision-making algorithms allow for the optimization of resource allocations in real-world applications such as taxi routing \cite{garces2023multiagent} and drone deliveries \cite{lee2022autonomous}. Optimal solution for these problems is intractable since it requires considering multiple scenarios of potential future requests and all relevant agent actions at each decision point, in extremely large state and control spaces that grow exponentially with the number of agents. Hence, many methods have been proposed to find competitive sub-optimal solutions. Most intuitive solutions are greedy policies, but they can be far from optimal--for example, when two taxis go to serve the same request, leaving other requests unattended. Studies \cite{kondor2022cost, vazifeh2018addressing} show that with non‐coordinating greedy policies, each additional ride‐hailing company in the market can largely increase the total number of circulating vehicles than necessary. Heuristic-based methods \cite{bertsekas1979distributed, bertsimas2019online, croes1958method} often generate myopic policies due to the lack of consideration for future demand. RL methods have been proposed in both offline \cite{ulmer2019offline, farazi2021deep} and online \cite{silver2010monte, somani2013despot, bent2004scenario} regimes; however, they either lack robustness to distribution shifts or are computationally expensive. Furthermore, sample efficiency remains a pervasive challenge in deep RL \cite{li2021breaking, li2023deep} due to sparse reward signals or large state spaces. As a result, deep Rl training from scratch typically requires millions of trajectories, which is expensive or impossible to collect in some cases. Given the substantial model sizes and pretraining data of LLMs, their rich world knowledge can enhance sample efficiency \cite{ahn2022can, morad2024language, ajay2024compositional, zhang2024improving} and robustness against distribution shift of the environment \cite{ge2024openagi, wang2023robustness}. For example, the previous art for this multi-agent routing problem \cite{garces2023multiagent} requires training separate a different network for each distinct representative demand distribution.

\subsection{Foundation models for Planning and Control}
The most common class of foundation models used for planning and control are vision-language-action models (VLAs) \cite{kim2024openvla, zhen20243d, brohan2023rt}, which leverage pretrained vision \cite{kim2024openvla,oquab2023dinov2}, language \cite{touvron2023llama,chowdhery2023palm}, or multimodal~\cite{zhai2023sigmoid,driess2023palm} models to interpret visual inputs, understand textual instructions, and generate contextually appropriate actions within dynamic environments. 

More relevant to us are those works using foundation models for spatial reasoning and path planning~\citep{yang2024diffusion, yang2024thinking, liu2023can}. \citeauthor{yang2024thinking} proposes a video-based benchmark to probe the spatial reasoning ability of multimodal LLMs (MLLMs). Their conclusions are rather negative: first, MLLMs are competitive but subhuman; second, linguistic prompting techniques are harmful for spatial reasoning. Instead of testing their ability in free-form environments, we take a step back, and test if they can reason spatially in an environment that can be represented as a graph. \citeauthor{liu2023can} applies Word2Vec models on real-world delivery route optimization \cite{merchan20242021} by drawing an analogy between delivery routes and sentences in language. Based on language descriptions of single/multi-robots routing tasks, \citeauthor{huang2024words} asks LLMs to generate Python code to solve single or multiple robots problems. Their evaluations are limited to prompting methods. They also perform task verification by generating unit tests. \citeauthor{deng2024can} investigates using LLMs as path planners and curriculum generators to mitigate hallucinations. The study leverages LLMs and Python to convert maze descriptions into Gym environments, where LLMs generate intermediate waypoints to simplify paths. Guided by LLMs, Q-learning iteratively plans paths, outputting the planning history and Q-table as the final policy.

The success of LLMs in multi-agent decision-making depends on their reasoning abilities, which are crucial for generating context-aware responses in dynamic environments. Techniques have been proposed for enhancing the reasoning capabilities of LLMs, including those based on CoT \cite{wei2022chain}, ToT \cite{yao2024tree}, Best-of-N \cite{lightman2023let, wang2023math}, Monte Carlo Tree Search (MCTS) \cite{gao2024interpretable, zhang2024rest, wang2023math, wang2024towards}, or search against learned verifiers \cite{cobbe2021training}. However, these approaches focus on single-agent scenarios and do not extend to multi-agent contexts, where the actions of one agent must be conditioned on the actions of others. Moreover, MCTS incurs significant computational costs during the search process \cite{wang2024q,ye2022spending}, severely restricts its applicability, especially in multi-agent scenarios.


\section{Limitations and Future Directions} 
The primary limitation of our work lies in the slower inference time of LLMs compared to GNN models or heuristic-based methods. 
In an online setting, this makes it impractical to use LLMs with a high number of MC samples, potentially hurting the performance.

As future work, We are especially interested in the following directions. First,  LLMs could be used to predict future requests or as value functions to evaluate generated plans. Second, while RL traditionally addresses environments with costly interactions, the compute expense of large-model inference may now surpass that of environmental interactions, necessitating new frameworks for this paradigm. Third, Diverse, challenging tasks are needed to test LLMs’ capabilities beyond traditional textual benchmarks, bridging the gap between current models and general robotics applications.


\section{Conclusion.} We have demonstrated that LLMs can plan effectively in structured spatial environments, even in zero-shot settings, when provided with well-crafted information and descriptions. Furthermore, we have shown that finetuning through rollout, which encourages models to consider multiple possible actions and future states, significantly enhances their performance. Based on prompting results, we conclude that adressing spatial hallucinations is crucial for further improving LLMs' spatial planning capabilities. We hope this work highlights the potential of leveraging pre-trained knowledge in LLMs to achieve greater efficiency in learning and executing complex multi-agent tasks. Additionally, we advocate for the development of more diverse and challenging multi-agent benchmarks to deepen our understanding of the capabilities embedded within these models.



\nocite{langley00}

\bibliography{paper}

\begin{thebibliography}{59}
\providecommand{\natexlab}[1]{#1}
\providecommand{\url}[1]{\texttt{#1}}
\expandafter\ifx\csname urlstyle\endcsname\relax
  \providecommand{\doi}[1]{doi: #1}\else
  \providecommand{\doi}{doi: \begingroup \urlstyle{rm}\Url}\fi

\bibitem[lla(2024)]{llamarecipes}
Llama cookbook: The official guide to building with llama models, 2024.
\newblock URL \url{https://github.com/meta-llama/llama-cookbook}.

\bibitem[Ahn et~al.(2022)Ahn, Brohan, Brown, Chebotar, Cortes, David, Finn, Fu, Gopalakrishnan, Hausman, et~al.]{ahn2022can}
Ahn, M., Brohan, A., Brown, N., Chebotar, Y., Cortes, O., David, B., Finn, C., Fu, C., Gopalakrishnan, K., Hausman, K., et~al.
\newblock Do as i can, not as i say: Grounding language in robotic affordances.
\newblock \emph{arXiv preprint arXiv:2204.01691}, 2022.

\bibitem[Ajay et~al.(2024)Ajay, Han, Du, Li, Gupta, Jaakkola, Tenenbaum, Kaelbling, Srivastava, and Agrawal]{ajay2024compositional}
Ajay, A., Han, S., Du, Y., Li, S., Gupta, A., Jaakkola, T., Tenenbaum, J., Kaelbling, L., Srivastava, A., and Agrawal, P.
\newblock Compositional foundation models for hierarchical planning.
\newblock \emph{Advances in Neural Information Processing Systems}, 36, 2024.

\bibitem[Bent \& Van~Hentenryck(2004)Bent and Van~Hentenryck]{bent2004scenario}
Bent, R.~W. and Van~Hentenryck, P.
\newblock Scenario-based planning for partially dynamic vehicle routing with stochastic customers.
\newblock \emph{Operations Research}, 52\penalty0 (6):\penalty0 977--987, 2004.

\bibitem[Bertsekas(2019)]{bertsekas2019reinforcement}
Bertsekas, D.
\newblock \emph{Reinforcement learning and optimal control}, volume~1.
\newblock Athena Scientific, 2019.

\bibitem[Bertsekas(2021)]{bertsekas2021multiagent}
Bertsekas, D.
\newblock Multiagent reinforcement learning: Rollout and policy iteration.
\newblock \emph{IEEE/CAA Journal of Automatica Sinica}, 8\penalty0 (2):\penalty0 249--272, 2021.

\bibitem[Bertsekas(2022)]{bertsekas2022lessons}
Bertsekas, D.
\newblock \emph{Lessons from AlphaZero for optimal, model predictive, and adaptive control}.
\newblock Athena Scientific, 2022.

\bibitem[Bertsekas(1979)]{bertsekas1979distributed}
Bertsekas, D.~P.
\newblock A distributed algorithm for the assignment problem.
\newblock \emph{Lab. for Information and Decision Systems Working Paper, MIT}, 1979.

\bibitem[Bertsekas \& Castanon(1999)Bertsekas and Castanon]{bertsekas1999rollout}
Bertsekas, D.~P. and Castanon, D.~A.
\newblock Rollout algorithms for stochastic scheduling problems.
\newblock \emph{Journal of Heuristics}, 5:\penalty0 89--108, 1999.

\bibitem[Bertsimas et~al.(2019)Bertsimas, Jaillet, and Martin]{bertsimas2019online}
Bertsimas, D., Jaillet, P., and Martin, S.
\newblock Online vehicle routing: The edge of optimization in large-scale applications.
\newblock \emph{Operations Research}, 67\penalty0 (1):\penalty0 143--162, 2019.

\bibitem[Brohan et~al.(2023)Brohan, Brown, Carbajal, Chebotar, Chen, Choromanski, Ding, Driess, Dubey, Finn, et~al.]{brohan2023rt}
Brohan, A., Brown, N., Carbajal, J., Chebotar, Y., Chen, X., Choromanski, K., Ding, T., Driess, D., Dubey, A., Finn, C., et~al.
\newblock Rt-2: Vision-language-action models transfer web knowledge to robotic control.
\newblock \emph{arXiv preprint arXiv:2307.15818}, 2023.

\bibitem[Chowdhery et~al.(2023)Chowdhery, Narang, Devlin, Bosma, Mishra, Roberts, Barham, Chung, Sutton, Gehrmann, et~al.]{chowdhery2023palm}
Chowdhery, A., Narang, S., Devlin, J., Bosma, M., Mishra, G., Roberts, A., Barham, P., Chung, H.~W., Sutton, C., Gehrmann, S., et~al.
\newblock Palm: Scaling language modeling with pathways.
\newblock \emph{Journal of Machine Learning Research}, 24\penalty0 (240):\penalty0 1--113, 2023.

\bibitem[Cobbe et~al.(2021)Cobbe, Kosaraju, Bavarian, Chen, Jun, Kaiser, Plappert, Tworek, Hilton, Nakano, et~al.]{cobbe2021training}
Cobbe, K., Kosaraju, V., Bavarian, M., Chen, M., Jun, H., Kaiser, L., Plappert, M., Tworek, J., Hilton, J., Nakano, R., et~al.
\newblock Training verifiers to solve math word problems.
\newblock \emph{arXiv preprint arXiv:2110.14168}, 2021.

\bibitem[Croes(1958)]{croes1958method}
Croes, G.~A.
\newblock A method for solving traveling-salesman problems.
\newblock \emph{Operations research}, 6\penalty0 (6):\penalty0 791--812, 1958.

\bibitem[Deng et~al.(2024)Deng, Zhang, Ou, and Feng]{deng2024can}
Deng, H., Zhang, H., Ou, J., and Feng, C.
\newblock Can llm be a good path planner based on prompt engineering? mitigating the hallucination for path planning.
\newblock \emph{arXiv preprint arXiv:2408.13184}, 2024.

\bibitem[Driess et~al.(2023)Driess, Xia, Sajjadi, Lynch, Chowdhery, Ichter, Wahid, Tompson, Vuong, Yu, et~al.]{driess2023palm}
Driess, D., Xia, F., Sajjadi, M.~S., Lynch, C., Chowdhery, A., Ichter, B., Wahid, A., Tompson, J., Vuong, Q., Yu, T., et~al.
\newblock Palm-e: An embodied multimodal language model.
\newblock \emph{arXiv preprint arXiv:2303.03378}, 2023.

\bibitem[Dubey et~al.(2024)Dubey, Jauhri, Pandey, Kadian, Al-Dahle, Letman, Mathur, Schelten, Yang, Fan, et~al.]{dubey2024llama}
Dubey, A., Jauhri, A., Pandey, A., Kadian, A., Al-Dahle, A., Letman, A., Mathur, A., Schelten, A., Yang, A., Fan, A., et~al.
\newblock The llama 3 herd of models.
\newblock \emph{arXiv preprint arXiv:2407.21783}, 2024.

\bibitem[Farazi et~al.(2021)Farazi, Zou, Ahamed, and Barua]{farazi2021deep}
Farazi, N.~P., Zou, B., Ahamed, T., and Barua, L.
\newblock Deep reinforcement learning in transportation research: A review.
\newblock \emph{Transportation research interdisciplinary perspectives}, 11:\penalty0 100425, 2021.

\bibitem[Gao et~al.(2024)Gao, Niu, He, Xu, Liu, Liu, Hu, and Wen]{gao2024interpretable}
Gao, Z., Niu, B., He, X., Xu, H., Liu, H., Liu, A., Hu, X., and Wen, L.
\newblock Interpretable contrastive monte carlo tree search reasoning.
\newblock \emph{arXiv preprint arXiv:2410.01707}, 2024.

\bibitem[Garces et~al.(2023)Garces, Bhattacharya, Gil, and Bertsekas]{garces2023multiagent}
Garces, D., Bhattacharya, S., Gil, S., and Bertsekas, D.
\newblock Multiagent reinforcement learning for autonomous routing and pickup problem with adaptation to variable demand.
\newblock In \emph{2023 IEEE International Conference on Robotics and Automation (ICRA)}, pp.\  3524--3531. IEEE, 2023.

\bibitem[Ge et~al.(2024)Ge, Hua, Mei, Tan, Xu, Li, Zhang, et~al.]{ge2024openagi}
Ge, Y., Hua, W., Mei, K., Tan, J., Xu, S., Li, Z., Zhang, Y., et~al.
\newblock Openagi: When llm meets domain experts.
\newblock \emph{Advances in Neural Information Processing Systems}, 36, 2024.

\bibitem[Gil et~al.(2023)Gil, Yemini, Chorti, Nedi{\'c}, Poor, and Goldsmith]{gil2023physicality}
Gil, S., Yemini, M., Chorti, A., Nedi{\'c}, A., Poor, H.~V., and Goldsmith, A.~J.
\newblock How physicality enables trust: A new era of trust-centered cyberphysical systems.
\newblock \emph{arXiv preprint arXiv:2311.07492}, 2023.

\bibitem[Huang et~al.(2024)Huang, Shi, and Sukhatme]{huang2024words}
Huang, Z., Shi, G., and Sukhatme, G.~S.
\newblock From words to routes: Applying large language models to vehicle routing.
\newblock \emph{arXiv preprint arXiv:2403.10795}, 2024.

\bibitem[Huggingface(2024)]{chat}
Huggingface.
\newblock chat templating, 2024.
\newblock URL \url{https://huggingface.co/docs/transformers/en/chat\_templating}.
\newblock Accessed: 2024-10-20.

\bibitem[Jumper et~al.(2021)Jumper, Evans, Pritzel, Green, Figurnov, Ronneberger, Tunyasuvunakool, Bates, {\v{Z}}{\'\i}dek, Potapenko, et~al.]{jumper2021highly}
Jumper, J., Evans, R., Pritzel, A., Green, T., Figurnov, M., Ronneberger, O., Tunyasuvunakool, K., Bates, R., {\v{Z}}{\'\i}dek, A., Potapenko, A., et~al.
\newblock Highly accurate protein structure prediction with alphafold.
\newblock \emph{nature}, 596\penalty0 (7873):\penalty0 583--589, 2021.

\bibitem[Kim et~al.(2024)Kim, Pertsch, Karamcheti, Xiao, Balakrishna, Nair, Rafailov, Foster, Lam, Sanketi, et~al.]{kim2024openvla}
Kim, M.~J., Pertsch, K., Karamcheti, S., Xiao, T., Balakrishna, A., Nair, S., Rafailov, R., Foster, E., Lam, G., Sanketi, P., et~al.
\newblock Openvla: An open-source vision-language-action model.
\newblock \emph{arXiv preprint arXiv:2406.09246}, 2024.

\bibitem[Kondor et~al.(2022)Kondor, Bojic, Resta, Duarte, Santi, and Ratti]{kondor2022cost}
Kondor, D., Bojic, I., Resta, G., Duarte, F., Santi, P., and Ratti, C.
\newblock The cost of non-coordination in urban on-demand mobility.
\newblock \emph{Scientific reports}, 12\penalty0 (1):\penalty0 4669, 2022.

\bibitem[Lee et~al.(2022)Lee, Shahzaad, Alkouz, Lakhdari, and Bouguettaya]{lee2022autonomous}
Lee, S., Shahzaad, B., Alkouz, B., Lakhdari, A., and Bouguettaya, A.
\newblock Autonomous delivery of multiple packages using single drone in urban airspace.
\newblock In \emph{Adjunct Proceedings of the 2022 ACM International Joint Conference on Pervasive and Ubiquitous Computing and the 2022 ACM International Symposium on Wearable Computers}, pp.\  72--74, 2022.

\bibitem[Li et~al.(2021)Li, Shi, Chen, Gu, and Chi]{li2021breaking}
Li, G., Shi, L., Chen, Y., Gu, Y., and Chi, Y.
\newblock Breaking the sample complexity barrier to regret-optimal model-free reinforcement learning.
\newblock \emph{Advances in Neural Information Processing Systems}, 34:\penalty0 17762--17776, 2021.

\bibitem[Li(2023)]{li2023deep}
Li, S.~E.
\newblock Deep reinforcement learning.
\newblock In \emph{Reinforcement learning for sequential decision and optimal control}, pp.\  365--402. Springer, 2023.

\bibitem[Lightman et~al.(2023)Lightman, Kosaraju, Burda, Edwards, Baker, Lee, Leike, Schulman, Sutskever, and Cobbe]{lightman2023let}
Lightman, H., Kosaraju, V., Burda, Y., Edwards, H., Baker, B., Lee, T., Leike, J., Schulman, J., Sutskever, I., and Cobbe, K.
\newblock Let's verify step by step.
\newblock \emph{arXiv preprint arXiv:2305.20050}, 2023.

\bibitem[Liu et~al.(2023)Liu, Wu, Liu, Wang, Wang, and Qu]{liu2023can}
Liu, Y., Wu, F., Liu, Z., Wang, K., Wang, F., and Qu, X.
\newblock Can language models be used for real-world urban-delivery route optimization?
\newblock \emph{The Innovation}, 4\penalty0 (6), 2023.

\bibitem[Merch{\'a}n et~al.(2024)Merch{\'a}n, Arora, Pachon, Konduri, Winkenbach, Parks, and Noszek]{merchan20242021}
Merch{\'a}n, D., Arora, J., Pachon, J., Konduri, K., Winkenbach, M., Parks, S., and Noszek, J.
\newblock 2021 amazon last mile routing research challenge: Data set.
\newblock \emph{Transportation Science}, 58\penalty0 (1):\penalty0 8--11, 2024.

\bibitem[Meta(2024{\natexlab{a}})]{llama3}
Meta, 2024{\natexlab{a}}.
\newblock URL \url{https://ai.meta.com/blog/meta-llama-3-1/}.
\newblock Accessed: 2024-12-15.

\bibitem[Meta(2024{\natexlab{b}})]{llama32}
Meta.
\newblock Introducing llama 3.2, 2024{\natexlab{b}}.
\newblock URL \url{https://www.llama.com/}.
\newblock Accessed: 2024-10-20.

\bibitem[Morad et~al.(2024)Morad, Shankar, Blumenkamp, and Prorok]{morad2024language}
Morad, S., Shankar, A., Blumenkamp, J., and Prorok, A.
\newblock Language-conditioned offline rl for multi-robot navigation.
\newblock \emph{arXiv preprint arXiv:2407.20164}, 2024.

\bibitem[OpenAI(2024)]{o1}
OpenAI.
\newblock Learning to reason with llms, 2024.
\newblock URL \url{https://openai.com/index/learning-to-reason-with-llms/}.
\newblock Accessed: 2024-12-15.

\bibitem[Oquab et~al.(2023)Oquab, Darcet, Moutakanni, Vo, Szafraniec, Khalidov, Fernandez, Haziza, Massa, El-Nouby, et~al.]{oquab2023dinov2}
Oquab, M., Darcet, T., Moutakanni, T., Vo, H., Szafraniec, M., Khalidov, V., Fernandez, P., Haziza, D., Massa, F., El-Nouby, A., et~al.
\newblock Dinov2: Learning robust visual features without supervision.
\newblock \emph{arXiv preprint arXiv:2304.07193}, 2023.

\bibitem[Silver \& Veness(2010)Silver and Veness]{silver2010monte}
Silver, D. and Veness, J.
\newblock Monte-carlo planning in large pomdps.
\newblock \emph{Advances in neural information processing systems}, 23, 2010.

\bibitem[Somani et~al.(2013)Somani, Ye, Hsu, and Lee]{somani2013despot}
Somani, A., Ye, N., Hsu, D., and Lee, W.~S.
\newblock Despot: Online pomdp planning with regularization.
\newblock \emph{Advances in neural information processing systems}, 26, 2013.

\bibitem[Touvron et~al.(2023)Touvron, Martin, Stone, Albert, Almahairi, Babaei, Bashlykov, Batra, Bhargava, Bhosale, et~al.]{touvron2023llama}
Touvron, H., Martin, L., Stone, K., Albert, P., Almahairi, A., Babaei, Y., Bashlykov, N., Batra, S., Bhargava, P., Bhosale, S., et~al.
\newblock Llama 2: Open foundation and fine-tuned chat models.
\newblock \emph{arXiv preprint arXiv:2307.09288}, 2023.

\bibitem[Ulmer et~al.(2019)Ulmer, Goodson, Mattfeld, and Hennig]{ulmer2019offline}
Ulmer, M.~W., Goodson, J.~C., Mattfeld, D.~C., and Hennig, M.
\newblock Offline--online approximate dynamic programming for dynamic vehicle routing with stochastic requests.
\newblock \emph{Transportation Science}, 53\penalty0 (1):\penalty0 185--202, 2019.

\bibitem[Vazifeh et~al.(2018)Vazifeh, Santi, Resta, Strogatz, and Ratti]{vazifeh2018addressing}
Vazifeh, M.~M., Santi, P., Resta, G., Strogatz, S.~H., and Ratti, C.
\newblock Addressing the minimum fleet problem in on-demand urban mobility.
\newblock \emph{Nature}, 557\penalty0 (7706):\penalty0 534--538, 2018.

\bibitem[Wang et~al.(2024{\natexlab{a}})Wang, Deng, Lv, Yan, and Bo]{wang2024q}
Wang, C., Deng, Y., Lv, Z., Yan, S., and Bo, A.
\newblock Q*: Improving multi-step reasoning for llms with deliberative planning.
\newblock \emph{arXiv preprint arXiv:2406.14283}, 2024{\natexlab{a}}.

\bibitem[Wang et~al.(2023{\natexlab{a}})Wang, Hu, Hou, Chen, Zheng, Wang, Yang, Huang, Ye, Geng, et~al.]{wang2023robustness}
Wang, J., Hu, X., Hou, W., Chen, H., Zheng, R., Wang, Y., Yang, L., Huang, H., Ye, W., Geng, X., et~al.
\newblock On the robustness of chatgpt: An adversarial and out-of-distribution perspective.
\newblock \emph{arXiv preprint arXiv:2302.12095}, 2023{\natexlab{a}}.

\bibitem[Wang et~al.(2023{\natexlab{b}})Wang, Li, Shao, Xu, Dai, Li, Chen, Wu, and Sui]{wang2023math}
Wang, P., Li, L., Shao, Z., Xu, R., Dai, D., Li, Y., Chen, D., Wu, Y., and Sui, Z.
\newblock Math-shepherd: A label-free step-by-step verifier for llms in mathematical reasoning.
\newblock \emph{arXiv preprint arXiv:2312.08935}, 2023{\natexlab{b}}.

\bibitem[Wang et~al.(2022)Wang, Wei, Schuurmans, Le, Chi, Narang, Chowdhery, and Zhou]{wang2022self}
Wang, X., Wei, J., Schuurmans, D., Le, Q., Chi, E., Narang, S., Chowdhery, A., and Zhou, D.
\newblock Self-consistency improves chain of thought reasoning in language models.
\newblock \emph{arXiv preprint arXiv:2203.11171}, 2022.

\bibitem[Wang et~al.(2024{\natexlab{b}})Wang, Song, Tian, Yu, Peng, Mi, Huang, and Yu]{wang2024towards}
Wang, X., Song, L., Tian, Y., Yu, D., Peng, B., Mi, H., Huang, F., and Yu, D.
\newblock Towards self-improvement of llms via mcts: Leveraging stepwise knowledge with curriculum preference learning.
\newblock \emph{arXiv preprint arXiv:2410.06508}, 2024{\natexlab{b}}.

\bibitem[Wei et~al.(2022)Wei, Wang, Schuurmans, Bosma, Xia, Chi, Le, Zhou, et~al.]{wei2022chain}
Wei, J., Wang, X., Schuurmans, D., Bosma, M., Xia, F., Chi, E., Le, Q.~V., Zhou, D., et~al.
\newblock Chain-of-thought prompting elicits reasoning in large language models.
\newblock \emph{Advances in neural information processing systems}, 35:\penalty0 24824--24837, 2022.

\bibitem[Yang et~al.(2024{\natexlab{a}})Yang, Su, Gkanatsios, Ke, Jain, Schneider, and Fragkiadaki]{yang2024diffusion}
Yang, B., Su, H., Gkanatsios, N., Ke, T.-W., Jain, A., Schneider, J., and Fragkiadaki, K.
\newblock Diffusion-es: Gradient-free planning with diffusion for autonomous and instruction-guided driving.
\newblock In \emph{Proceedings of the IEEE/CVF Conference on Computer Vision and Pattern Recognition}, pp.\  15342--15353, 2024{\natexlab{a}}.

\bibitem[Yang et~al.(2024{\natexlab{b}})Yang, Yang, Gupta, Han, Fei-Fei, and Xie]{yang2024thinking}
Yang, J., Yang, S., Gupta, A.~W., Han, R., Fei-Fei, L., and Xie, S.
\newblock Thinking in space: How multimodal large language models see, remember, and recall spaces.
\newblock \emph{arXiv preprint arXiv:2412.14171}, 2024{\natexlab{b}}.

\bibitem[Yang et~al.(2023)Yang, Nachum, Du, Wei, Abbeel, and Schuurmans]{yang2023foundation}
Yang, S., Nachum, O., Du, Y., Wei, J., Abbeel, P., and Schuurmans, D.
\newblock Foundation models for decision making: Problems, methods, and opportunities.
\newblock \emph{arXiv preprint arXiv:2303.04129}, 2023.

\bibitem[Yao et~al.(2024)Yao, Yu, Zhao, Shafran, Griffiths, Cao, and Narasimhan]{yao2024tree}
Yao, S., Yu, D., Zhao, J., Shafran, I., Griffiths, T., Cao, Y., and Narasimhan, K.
\newblock Tree of thoughts: Deliberate problem solving with large language models.
\newblock \emph{Advances in Neural Information Processing Systems}, 36, 2024.

\bibitem[Ye et~al.(2022)Ye, Abbeel, and Gao]{ye2022spending}
Ye, W., Abbeel, P., and Gao, Y.
\newblock Spending thinking time wisely: Accelerating mcts with virtual expansions.
\newblock \emph{Advances in Neural Information Processing Systems}, 35:\penalty0 12211--12224, 2022.

\bibitem[Zhai et~al.(2023)Zhai, Mustafa, Kolesnikov, and Beyer]{zhai2023sigmoid}
Zhai, X., Mustafa, B., Kolesnikov, A., and Beyer, L.
\newblock Sigmoid loss for language image pre-training.
\newblock In \emph{Proceedings of the IEEE/CVF International Conference on Computer Vision}, pp.\  11975--11986, 2023.

\bibitem[Zhang et~al.(2024{\natexlab{a}})Zhang, Zhoubian, Yue, Dong, and Tang]{zhang2024rest}
Zhang, D., Zhoubian, S., Yue, Y., Dong, Y., and Tang, J.
\newblock Rest-mcts*: Llm self-training via process reward guided tree search.
\newblock \emph{arXiv preprint arXiv:2406.03816}, 2024{\natexlab{a}}.

\bibitem[Zhang et~al.(2024{\natexlab{b}})Zhang, Zhu, Saphra, Kleiman, Edelman, Tambe, Kakade, and Malach]{zhang2024transcendence}
Zhang, E., Zhu, V., Saphra, N., Kleiman, A., Edelman, B.~L., Tambe, M., Kakade, S.~M., and Malach, E.
\newblock Transcendence: Generative models can outperform the experts that train them.
\newblock \emph{arXiv preprint arXiv:2406.11741}, 2024{\natexlab{b}}.

\bibitem[Zhang et~al.(2024{\natexlab{c}})Zhang, Li, Li, Zhang, Yu, and Ye]{zhang2024improving}
Zhang, F., Li, J., Li, Y.-C., Zhang, Z., Yu, Y., and Ye, D.
\newblock Improving sample efficiency of reinforcement learning with background knowledge from large language models.
\newblock \emph{arXiv preprint arXiv:2407.03964}, 2024{\natexlab{c}}.

\bibitem[Zhen et~al.(2024)Zhen, Qiu, Chen, Yang, Yan, Du, Hong, and Gan]{zhen20243d}
Zhen, H., Qiu, X., Chen, P., Yang, J., Yan, X., Du, Y., Hong, Y., and Gan, C.
\newblock 3d-vla: A 3d vision-language-action generative world model.
\newblock \emph{arXiv preprint arXiv:2403.09631}, 2024.

\end{thebibliography}
\bibliographystyle{icml2025}

\section*{Impact Statement}
This paper presents work whose goal is to advance the field of Machine Learning. There are many potential societal consequences of our work, none which we feel must be specifically highlighted here.

\newpage
\appendix
\onecolumn


\section{Bar Charts for Standard Deviation}
Within each load level, the difficulty of scenarios can vary significantly depending on when and where requests occur. Therefore, we present error bars representing the standard deviation of costs in the $400\times400 m^2$ map across all scenarios within each load level to illustrate this variability.
\vskip -0.5in
\begin{figure}[ht]
\vskip -0.1in
\begin{center}
\centerline{\includegraphics[height=2.2in]{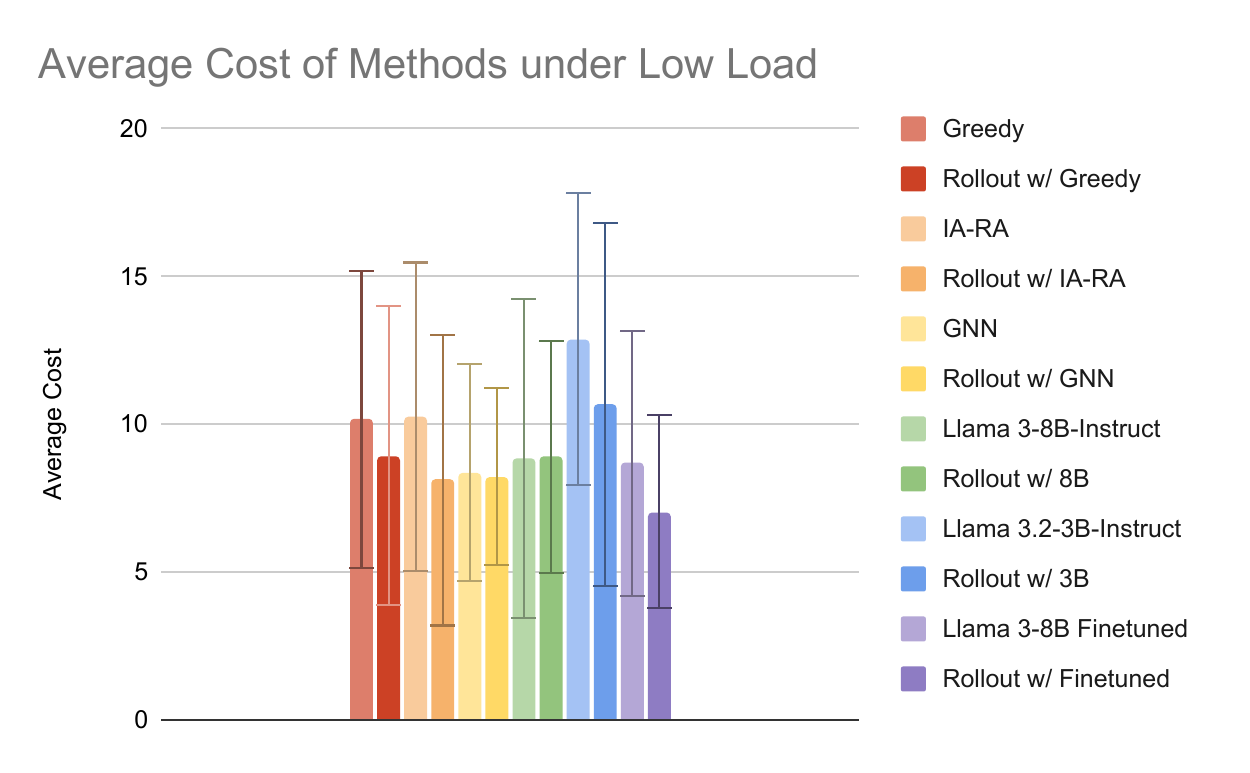}}
\caption{Average cost and error bars in terms of standard deviation of methods over test set under the low load level. }
\label{sample-complexity}
\end{center}
\vskip -0.1in
\end{figure}

\vskip -1.1in

\begin{figure}[ht]
\vskip -0.2in
\begin{center}
\centerline{\includegraphics[height=2.2in]{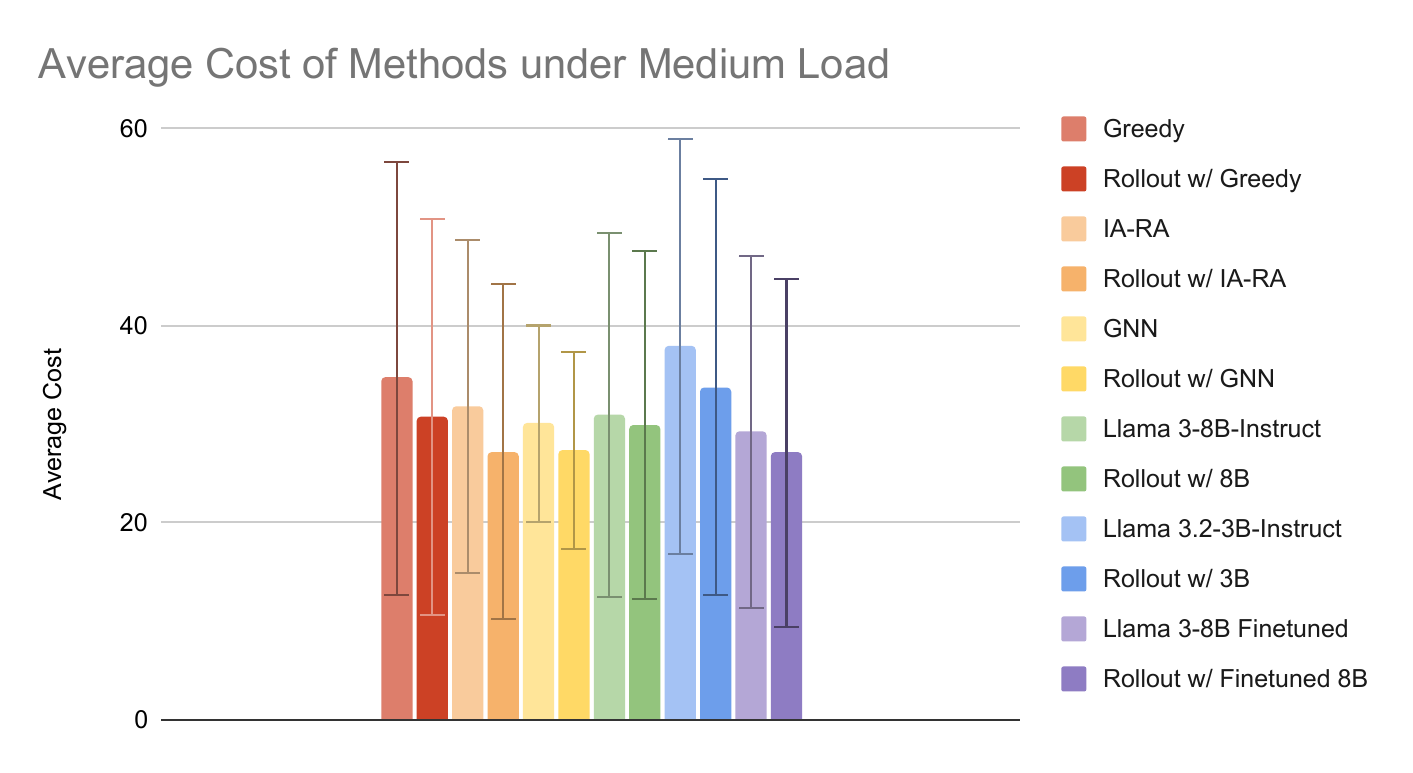}}
\caption{Average cost and error bars in terms of standard deviation  of methods over test set under the medium load level.}
\label{sample-complexity}
\end{center}
\vskip -1in
\end{figure}

\vskip -0.1in

\begin{figure}[ht]
\vskip -0.2in
\begin{center}
\centerline{\includegraphics[height=2.2in]{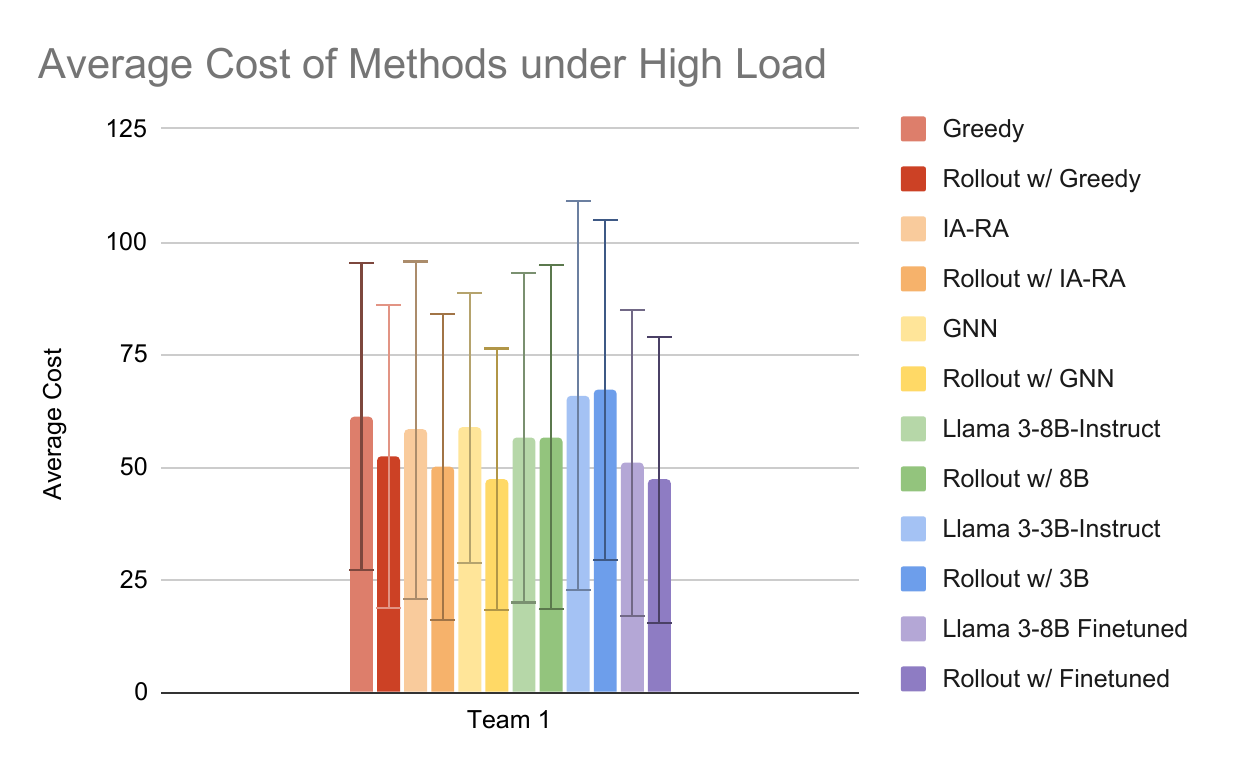}}
\caption{Average cost and error bars in terms of standard deviation  of methods over test set under the high load level.}
\label{sample-complexity}
\end{center}
\vskip -0.1in
\end{figure}
 

\section{Prompts}
\label{sec: prompts}

\subsection{System and User Prompts}
\label{sec: sys_prompt}
\begin{mytextbox}{System and User Prompts as State}
\textbf{System:}  "You are a taxi driver in a multi-taxicab team on a map described by roads and intersections. Nodes (intersections) are listed by index with coordinates (longitude, latitude): $65293741$: (-122.4097034, 37.7817636), 65293743: (-122.4092587, 37.7814105), 65303533: (-122.4038718, 37.7898332), $\dots$

Roads are expressed as connections between nodes in the form 'from node to node': 65293741 to 65293743, 65293741 to 65318282, 65293741 to 1723738829, 65293743 to 65293741, 65293743 to 65306931, 65303533 to 65303538, 65303538 to 6378899319, 65303538 to 1271001348, 65303541 to 65303544, 65303541 to 1271001343, $\dots$

Your goal is to minimize the waiting time of all riders. Make your decisions based on the following rules:

1. A taxi can only pick up an active request if it is idle.

2. If no request exists or you choose not to pick up, you must decide where to move next."
\vspace{1em}

\textbf{User:} You are Taxi 0. You may only pick up a request if there is active request in the system. Taxi 0 is idle at location 6925582021. Taxi 1 is idle at location 1578907668. Taxi 2 is idle at location 65306810. Next action for taxi 1: do not pickup, go to 1578907668. Next action for taxi 2: do not pickup, go to 65306810. Please provide your next action as a tuple in the format: (pickup: True or False, next position: [numeric value]).
\vspace{1em}

\textbf{Assistant:} As Taxi 0, I am idle at location 6925582021. I will check for active requests.

There are no active requests in the system. Therefore, I will decide where to move next. I will move to the closest idle taxi to reduce waiting time.

The closest idle taxi is Taxi 2, which is at location 65306810. I will move to location $65306810$.

My next action is: (pickup: False, next position: 65306810)

\end{mytextbox}
\subsection{Chain-of-Thought}
\label{sec: cot_prompt}
\begin{mytextbox}{Chain-of-Thought}
\textbf{User:} [Zero-shot User Prompt as in \cref{sec: sys_prompt}]

Before deciding your next action, consider the following steps: \\
1. Identify all outstanding requests.\\
2. Determine which request you should prioritize picking up, aiming to minimize the total waiting time. 
 
  \quad - You should pick up a request immediately if you are at the same location/one step away from the request. 
  
  \quad - You can move to a node and pick up the request at that node in the same step.

3. Evaluate which requests other agents are already closer to or actively moving toward, based on their known or expected actions. Adjust your action if necessary.\\
After reasoning through these steps, provide your next action as a tuple in the format: (pickup: True or False, next position: a numeric value).\\
Remember, your goal is to minimize total waiting time and avoid targeting requests that are better suited for other agents unless no alternative exists.

\vspace{1em}

\end{mytextbox}

\subsection{Tree-of-Thoughts}
\label{sec: tot_prompt}
\begin{mytextbox}{Tree-of-Thoughts}
\textbf{System:} [Zero-shot User Prompt as in \cref{sec: sys_prompt}]

Before deciding your next action, think through the following tree of possibilities:\\
1.	Identify Requests:

    \quad - Which requests are outstanding?
    
	\quad - Which are closest or most urgent?
    
	\quad - Are other agents closer or already handling them?\\
→ Rank requests by priority based on proximity, urgency, and agent competition.\\
2.	Evaluate Actions:

	\quad - Action 1: Pick up a request if you're at/one step away. You can move to a node and pick up the request at that node in the same step.
    
	\quad - Action 2: Move toward a high-priority request.
    
	\quad - Action 3: Stay if no better option exists.\\
→ Simulate the impact of each action on total waiting time.

3.	Decide:

    \quad - Which action minimizes waiting time?
    
    \quad - Does it avoid unnecessary conflicts with other agents?\\
→ Choose the best action as: (pickup: True/False, next position: numeric value).




\end{mytextbox}

\vskip -1em
\subsection{Few-shot with CoT}
\begin{mytextbox}{Few-Shot}
\textbf{User:} Here are three examples: 
$\langle|$start\_header\_id$|\rangle$user$\langle|$end\_header\_id$|\rangle$ You are Taxi 0. You are idle at location 65334120. Taxi 1 is idle at location 1580501206. Taxi 2 is idle at location 1580501206. Currently there are outstanding requests in the system:

pickup\_location: 65314158

  \quad Taxi 0 shortest route: $[65334120, 65314158]$ (length: 1)
  
  \quad Taxi 1 shortest route: [1580501206, 65334120, 65314158] (length: 2)
  
  \quad Taxi 2 shortest route: [1580501206, 65334120, 65314158] (length: 2)
  
 Expected next action for taxi 1: go to 65334120. Expected next action for taxi 2: go to 65334120. You should prefer picking up requests closer to you. You should not follow other agents. Please provide your next action as a tuple in the format: (pickup: True or False, next position: a numeric value).
 
$\langle|$start\_header\_id$|>$assistant$\langle|$end\_header\_id$|>$ I'm the closest to the request at 65314158 among all taxis. I will pick it up. My next action is: (pickup: True, next position: 65314158). \\

$\langle|$start\_header\_id$|\rangle$user$\langle|$end\_header\_id$|\rangle$ You are Taxi 1. Taxi 0 is idle at location 65314156. You are idle at location 65317939. Taxi 2 is idle at location 65314156. Currently there are outstanding requests in the system:

  pickup\_location: 1580501214
  
    \quad Taxi 1 shortest route: [65317939, 1580501214] (length: 2)
    
    \quad Taxi 0 shortest route: [65314156, 6988532585, 386885670, 1271001343, 6988532615, 2936165726, 65317939, 1580501214] (length: 8)
    
    \quad Taxi 2 shortest route: [65314156, 6988532585, 386885670, 1271001343, 6988532615, 2936165726, 65317939, 1580501214] (length: 8)
    
  Known next action for taxi 0: go to 65314156. Expected next action for taxi 2: go to 6988532585. Provide your next action as a tuple in the format: (pickup: True or False, next position: a numeric value).
  
$\langle|$start\_header\_id$|\rangle$assistant$\langle|$end\_header\_id$|\rangle$ I'm the closest to the request at 1580501214 among all taxis. I will pick it up. My next action is: (pickup: True, next position: 1580501214). \\

$\langle|$start\_header\_id$|\rangle$user$\langle|$end\_header\_id$|\rangle$ You are Taxi 1.  Taxi 0 is idle at location 1308305528. You are idle at location 552853360.  Taxi 2 is idle at location 65313133.  Currently there are outstanding requests in the system:

  pickup\_location: 6988532585
  
    \quad Taxi 0 shortest route: [1308305528, 6988532585] (length: 1)
    
    \quad Taxi 1 shortest route: [552853360, 1308305528, 6988532585] (length: 2)
    
    \quad Taxi 2 shortest route: [65313133, 65313138, 1578907668, 552853360, 1308305528, 6988532585] (length: 5)
    
  Known next action for taxi 0: go to 6988532585. Expected next action for taxi 2: go to 65313138. You should prefer picking up requests closer to you. You should not follow other agents. Please provide your next action as a tuple in the format: (pickup: True or False, next position: a numeric value).
  
$\langle|$start\_header\_id$|\rangle$assistant$\langle|$end\_header\_id$|\rangle$ Since there is only one request at 6988532585 and Taxi 0 is closer to it and trying to pick it up, I will not try to pick it up. My next action is: (pickup: False, next position: 552853360).\\

Now, [Zero-shot User Prompt as in \cref{sec: sys_prompt}]


\end{mytextbox}

\section{Examples of Hallucination}
\label{sec: hallu}
\begin{mytextbox}{Hallucination}


\textbf{Assistant:} To minimize the total waiting time of all riders, I will follow the given constraints and prioritize picking up requests closer to my current location.

My current location is 65328690. 

The shortest route to the first request 65343958 is [65328690, 2936165726, 65317939, 65371286, 65332806, 65313133, 65326742, 65326744, 65343958]. 

The shortest route to the second request 65303546 is [65328690, 2936165726, 3902413693, 552853360, 1308305528, 6988532585, 65303544, 65303546]. 

Since the first request 65343958 is closer to me, I will pick it up. The second request 65303546 is farther away, so I will not pick it up.

My next action is to pick up the request at location 65343958. (pickup: True, next position: 65343958)

\end{mytextbox}

In this example, the LLM successfully identified the shortest paths to the two outstanding requests by itself. However, it mistakenly concluded that the first request was closer, while the second was actually nearer—a case of \textbf{spatial} hallucination. Furthermore, despite heading toward the first request, it should have chosen Node 2936165726 as the next step. Instead, it exhibited a \textbf{reachability} hallucination by selecting Node 65343958.

\section{Examples of Additional Semantic Information}
\label{sec: semantic-info}
\begin{mytextbox}{Semantic Information: Conditioning on Weather}

\textbf{User:}  … Current weather in San Francisco: heavy rain with temperature: 11.8 ℃ and wind speed: 7.9 km/h.   
 Please provide your next action as a tuple in the format: (pickup: True or False, next position: a numeric value). 

\vspace{1em}

\textbf{Assistant:} Given the current weather conditions, we can assume that the travel time will be increased by 1 unit due to the heavy rain…

\end{mytextbox}


\end{document}